\pdfoutput=1

\documentclass[11pt]{article}

\usepackage[final]{coling}

\usepackage{times}
\usepackage{latexsym}

\usepackage[T1]{fontenc}

\usepackage[utf8]{inputenc}

\usepackage{microtype}

\usepackage{inconsolata}

\usepackage{graphicx}

\usepackage{amsmath, amssymb}
\usepackage{multirow, multicol}
\usepackage{booktabs, graphicx}
\usepackage{subfigure}
\usepackage{marvosym}
\usepackage[linesnumbered,ruled,vlined]{algorithm2e}
\usepackage{booktabs}
\usepackage{hyperref}
\usepackage{multirow, multicol, makecell}
\newcommand{\nop}[1]{}

\title{Mitigating Out-of-Entity Errors in Named Entity Recognition: \\ A Sentence-Level Strategy}

\author{Guochao Jiang$^{1\dag}$, Ziqin Luo$^{1\dag}$, Chengwei Hu$^{1\ddag}$, Zepeng Ding$^{1\dag}$, Deqing Yang$^{12\ddag\textrm{\Letter}}$ \\
$^1$School of Data Science, Fudan University, Shanghai, China \\
$^2$Shanghai Key Laboratory of Data Science, Shanghai, China \\
$^\dag$\texttt{\{gcjiang22,zqluo22,zpding22\}@m.fudan.edu.cn} \\
$^\ddag$\texttt{\{cwhu20,yangdeqing\}@fudan.edu.cn}\\
}

\begin{document}

\maketitle

\begin{abstract}
Many previous models of named entity recognition (NER) suffer from the problem of Out-of-Entity (OOE), i.e., the tokens in the entity mentions of the test samples have not appeared in the training samples, which hinders the achievement of satisfactory performance. To improve OOE-NER performance, in this paper, we propose a new framework, namely S+NER, which fully leverages sentence-level information. Our S+NER achieves better OOE-NER performance mainly due to the following two particular designs. 1) It first exploits the pre-trained language model's capability of understanding the target entity's sentence-level context with a template set. 2) Then, it refines the sentence-level representation based on the positive and negative templates, through a contrastive learning strategy and template pooling method, to obtain better NER results. Our extensive experiments on five benchmark datasets have demonstrated that, our S+NER outperforms some state-of-the-art OOE-NER models.
\end{abstract}


\section{Introduction}\label{sec:intro}
Named entity recognition (NER) \cite{NER} plays a vital role in many downstream tasks including knowledge graph construction \cite{CNDB}, information retrieval \cite{banerjee2019information}, question answering \cite{molla2006named}, etc. NER task aims to recognize the span(s) of entity mention(s) in one input sentence, and further type the entity(s) mentioned by the span(s). For example, given a sentence ``Aurora Couture aims to illuminate the fashion industry with its unique designs and commitment to ethical practices'', the model should output `Aurora Couture' as the entity mentioned and type the entity as `Brand'.

\begin{figure}
    \centering
    \includegraphics[width=0.47 \textwidth]{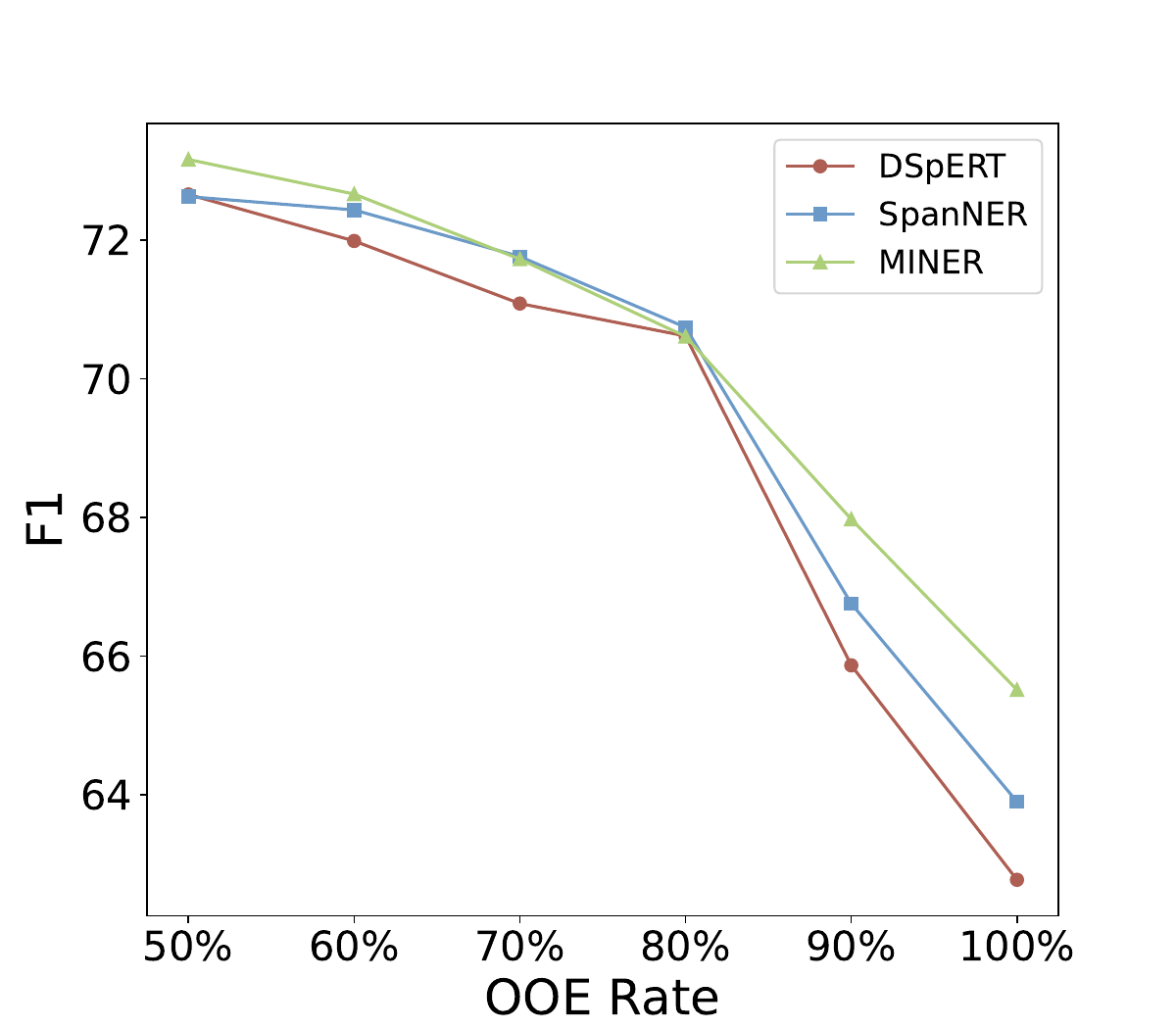}
    \caption{The span-based NER models' F1 scores on TwitterNER dataset with different OOE rates. The OOE rate is defined as the ratio of the test entities whose mention spans have the words (tokens) not appearing in the training set, to all entities in the test set. The numbers here are the results of deduplication, and duplicate entities are not considered.
    }
    \label{fig:OOE_rate}
\end{figure}

In recent years, many deep NER models \cite{ma2016end,zhu2018gram,wang2020astral,wang2022miner,deepspan,toner} have exhibited state-of-the-art (SOTA) performance. 
Nevertheless, these models' performance declines when many words (tokens) in the entity mentions have not appeared in the training set before, which is referred to as \emph{Out-of-Entity} (OOE) problem \cite{mikheev1999named}. As shown in Figure \ref{fig:OOE_rate}, the F1 score of the representative models (SpanNER \cite{fu-etal-2021-spanner}, MIENR \cite{wang2022miner} and DSpERT \cite{deepspan}) drops apparently as the OOE rate increases. It is probably because the span representations of OOE words have not been fine-tuned during model training, while most of the previous models achieve NER mainly based on the span representations. 

The previous NER models achieve the NER of OOE (OOE-NER) mainly through leveraging external knowledge, OOE word embedding, or contextualized embedding. The methods of external knowledge \cite{zhang-yang-2018-chinese,li-etal-2018-self,P-ICL,agentre} enhance their recognition capabilities by incorporating external information. However, external knowledge especially domain-specific knowledge, may not be obtained conveniently and cheaply. The methods of OOE word embedding \cite{fukuda-etal-2020-robust, peng2019learning} strive to construct more robust embeddings for OOE words, but they only leverage the semantics of the OOE words themselves without considering other useful information such as sentence-level context. Comparatively, the methods of contextualized embedding \cite{hu-etal-2019-shot} incorporate the contextual information of the mentions besides the OOE word embeddings to achieve NER, through learning the appropriate embeddings of the tokens near the OOE words. 
Nonetheless, these models have not fully exploited the \emph{sentence-level} contextual information of OOE words which is indeed helpful to improve OOE-NER performance further.

Intuitively, the semantic and syntactic information in the context of a new entity's mention can prompt us to recognize the entity's name and identify its type correctly. Recall the sentence illustrated in the first paragraph, 
when `Aurora Couture' is replaced with another span, we still believe the new span is also the name of a brand or company. Another example is ``XXX is a wonderful city.'', where the type of `XXX' can be recognized as location easily through understanding the whole sentence. Inspired by it, we propose a new NER framework, namely \emph{S+NER}, which fully leverages the mention's contextual information to enhance OOE-NER performance. Since the OOE words in the target mention have not appeared in the training samples resulting in unsatisfactory OOE embeddings, our S+NER additionally incorporates the representation of the whole sentence, which is generated by a BERT-based encoder \cite{devlin2018bert}. Furthermore, S+NER refines the sentence representation by adopting a contrastive learning strategy. As we mentioned before, contextual information is highly correlated with the correct understanding of the entity and its type. As a result, an optimal context representation in terms of accurate NER should be close to the representation of the positive template involving the entity span and its correct type, and simultaneously far away from the negative templates involving the entity span and its incorrect types. That is the basic principle of our proposed contrastive learning strategy.

Our main contributions in this paper are summarized as follows.

\begin{enumerate}
    \item We propose a novel framework S+NER based an effective contrastive learning method that fully leverages the sentence-level information of the mention(s) to achieve improved OOE-NER performance. 
    \item We utilize the large language model (LLM) to generate a template set for OOE-NER that is both high-quality and semantically rich. To enhance performance, we incorporate sentence-level information from various templates using a pooling method designed for multiple templates.
    \item We conduct extensive experiments on some benchmark datasets to justify S+NER's advantage on OOE-NER over some SOTA OOE-NER models.
\end{enumerate}

The rest of our paper is organized as follows. We briefly introduce the research works related to our work in Section 2 and detail our proposed framework including its special designed components in Section 3. Then, we display our extensive experiment results and provide in-depth analysis in Section 4. At last, we conclude our work in Section 5.
\section{Related Work}

\subsection{NER Solutions for OOE Problem}
By now, in the research field of NER, the OOE problem has been addressed through three primary categories of approaches as follows.

The first category incorporates external knowledge to tackle the OOE issue. \citet{zhang-yang-2018-chinese} exemplify this by constructing extensive entity dictionaries, thereby enhancing the model's look-up capability. However, this approach often compromises the model's generalization ability. \citet{li-etal-2018-self} attempted to mitigate this by introducing part-of-speech tags as external knowledge. However, this method is contingent on the availability of such tags and a high-quality external knowledge base, which is often challenging to procure.

The second category focuses on enhancing OOE word embeddings. For instance, \citet{10.1162/tacl_a_00051} utilized each word's character-level n-gram to represent the OOE word embedding, given the absence of OOE words at the character level. \citet{pinter-etal-2017-mimicking} employed character-level Recurrent Neural Networks (RNNs) to capture morphological features. Other methods involve using known words during the training process to match OOE words, subsequently replacing the OOE word embeddings with the embeddings of the matched known words. \citet{peng2019learning} trained a student network to predict the closest word embedding for OOE words in the representation space. \citet{fukuda-etal-2020-robust} leveraged the known words similar in character surface form to the OOE words. However, these methods typically yield a static OOE embedding, neglecting the exploitation of contextual information.

The third category leverages contextual information to enhance the representation of OOE words. \citet{hu-etal-2019-shot} reformulated the OOE problem as a K-shot regression problem, predicting OOE embedding by aggregating K context features. Recently, researchers have utilized pre-trained language models (PLMs) \cite{peters-etal-2018-deep, devlin2018bert, liu2019roberta, he2020deberta}, given their demonstrated proficiency in embedding contextual words with contextual information. However, \citet{yan-etal-2021-unified-generative} found that BERT does not always outperform BiLSTM-CRF in capturing contextual information. \citet{toner} introduced ToNER to solve NER by using an entity type matching model to identify potential entity types in the sentence and leverage contextual information. In contrast to these methods, our proposed S+NER model leverages sentence-level contextual information, as opposed to word-level and character-level embeddings.

\subsection{NER Solutions with Contrastive Learning}
Contrastive Learning is effective in representation learning practice, which is commonly aimed at improving the alignment and uniformity of the representation space to optimize the learned representations~\cite{20-cl-alignment-uniformity}.
Therefore, contrastive learning is widely applied in NER to improve entity representations thus further improving model performance.
ERICA~\cite{21-erica} introduces an entity-level pre-training task called entity discrimination, which optimizes entity representation by contrasting the semantics of different types of entities.
CONTaiNER~\cite{22-container} introduces contrastive learning at the token-level to reduce the distance between token embeddings of the same entity type and increase that between different types, alleviating the overfitting problem during downstream task transfer.
BINDER~\cite{23-binder} contrasts the entity type description with all the tokens in the input text from both the entity and token perspectives, to enhance nested and discontinuous entity extraction. 
Similar ideas are also found in ToNER~\cite{toner}.
CLLMFS~\cite{24-cllmfs} incorporates contrastive loss into the instruction fine-tuning phase to enhance the LLM's understanding of entity mentions and entity types.
In this work, we utilize contrastive learning to optimize the model's comprehension of sentence-level contextual information, thereby improving OOE-NER performance.

\section{Methodology}
In this section, we first introduce the span classification task for NER (in Section \ref{sec:SpanNER}). Then, we introduce the model design in S+NER (in Section \ref{sec:S+NER}), which are used to obtain optimal context representation for enhanced OOE-NER.

\subsection{Backbone Model}\label{sec:SpanNER}
In recent years, the model architecture for NER with pre-trained language models (PLMs) has changed from the initial sequence labeling task \cite{10.1162/tacl_a_00104, akbik-etal-2018-contextual, DBLP:journals/corr/abs-1911-04474} to the span classification task \cite{mengge-etal-2020-coarse, yamada-etal-2020-luke, fu-etal-2021-spanner, deepspan}. 
We choose span-based tasks for NER as the backbone of our S+NER for the following two reasons.

\begin{enumerate}
    \item Span-based model can extract the explicit information of entity mention spans, which can be easily leveraged by an advanced NER model \cite{deepspan}. 
    \item Compared with other sequence labeling models of NER, Span-based model has demonstrated better generalization capability on OOE scenarios \cite{fu-etal-2021-spanner}.
\end{enumerate}

The classical span-based model for NER mainly consists of three parts: token representation layer, span representation layer, and span classification layer, which are introduced in turn as follows.

\subsubsection{Token Representation Layer}

Formally, let $X = \{x_1, x_2, \cdots, x_n\}$ represent the input sentence sequence, $\mathbf{h}_i$ denote the representation or hidden state of the $i$-th token $x_i$ respectively. In most cases, they are as follows:
\begin{align}
\mathbf{h}_1, \mathbf{h}_2, \cdots, \mathbf{h}_n &= \text{Encoder}(x_1, x_2, \cdots, x_n),
\end{align}
where $\text{Encoder}(\cdot)$ can be implemented with any network structure with context encoding function, e.g., LSTM \cite{6795963}, Transformer \cite{NIPS2017_3f5ee243} and so on. In most cases, $\mathbf{h}_i \in \mathbb{R}^d, 1 \le i \le n$, which means that the dimension of token embedding vector is equal to the dimension of token representation vector for PLMs. $d$ is the dimension of the PLM, e.g. $d=768$ for BERT-base and $d=1024$ for BERT-large.

\subsubsection{Span Representation Layer}

Denote $S=\{s_1, s_2, ..., s_m\}$ is the set of all potential spans in sequence $X$ where $b_i$ and $e_i$ are the start position index and the end position index of span $s_i = (b_i, e_i), 1\leq i\leq m$, respectively. For each span $s_i$, a label $y_i$ is assigned to indicate whether $s_i$ is the entity span of a certain type or non-entity span (denoted by O in most NER tasks). For example, given a sentence $X$ = ``Milan is wonderful.'', its potential span set is $S = \{(1,1),(1,2),(1,3),(2,2),(2,3),(3,3)\}$, and the corresponding label set is $\{\text{LOC},\text{O},\text{O},\text{O},\text{O},\text{O}\}$ where LOC indicates the span `Milan' is a location entity.

For the span $s_i=(b_i,e_i)$, its representation consists of two parts in the classical span-based NER model: \textbf{boundary embedding} and \textbf{span length embedding}.

\begin{itemize}
    \item \textbf{Boundary Embedding}: For the NER model based on span classification, the text dimension information of the span itself must be the key to downstream tasks. One method is to use the representation of each token in a span and concatenate them together as the information representation of this part. However, this method will produce representations of different dimensions for spans of different lengths, and for spans that are too long, the representation dimension will be too large. The classic method is to use the representation of two boundary tokens of span. Formally, denote $\mathbf{z}_i^b$ as the boundary embedding. It is the concatenation of the start token's representation $\mathbf{h}_{b_i}$ and the end token's representation $\mathbf{h}_{e_i}$.
\begin{align}
    \mathbf{z}_i^b =[\mathbf{h}_{b_i}; \mathbf{h}_{e_i}]\in\mathbb{R}^{2d}.
\end{align}
    \item \textbf{Span Length Embedding}: If only the boundary embedding discussed above is used as a span representation for subsequent entity classification, this method will lose the explicit positional information, even if the Transformer architecture has its own positional encoding in most cases \cite{learned_positional_embedding, rope}. At the same time, due to the use of boundary embedding, the length of the span itself cannot be used. Therefore, in general methods, an additional learnable span length embedding is added as part of the span representation. Formally, denote $\mathbf{z}_i^l \in \mathbb{R}^{d'}$ as the span length embedding, which is initialized randomly and learned through model training.
\end{itemize}

Then, denote $\mathbf{z}_i$ as the span $s_i$'s representation, which is the concatenation of the boundary embedding and span length embedding:
\begin{align}
    \mathbf{z}_i = [\mathbf{z}_i^b; \mathbf{z}_i^l]\in\mathbb{R}^{2d+d'}.
\end{align}

\subsubsection{Span Classification Layer}

Upon acquiring the span representation $\mathbf{z}_i$, it is fed into a fully connected neural network. The purpose of this operation is to classify the span in accordance with the entity type label that the model suggests for the span $s_i$. Formally, we denote the span classifier as a function $F$. This function $F$ maps a span representation of dimension $2d + d'$ to a vector of dimension $|\mathcal{Y}|$, where $\mathcal{Y}$ represents the entity type set. The resultant vector $F(\mathbf{z}_i)$ embodies scores that correspond to each unique label for the span $s_i$. The model's predicted label $\hat{y}$ for the span $s_i$ is determined by identifying the label that corresponds to the highest score in the vector.


For the overlapped spans, only the span with the largest score is reserved at last. Using this heuristic decoding method to achieve flat NER \cite{NER} can avoid predicting overlapped spans.

\subsubsection{Learning Objective}

Suppose $y_i$ is the true label of $s_i$, the loss function with respect to a training sample (sentence) in span-based NER model is
\begin{equation}
    \mathcal{L}_1 = -\sum\limits_{s_i\in S} \mathrm{CE}(F(\mathbf{z}_i), y_i),
\end{equation}
where $S$ is the sentence's span set from which the overlapped spans have been removed, and $\mathrm{CE}(\cdot, \cdot)$ is the cross entropy function for score vector $F(\mathbf{z}_i)$ and corresponding true label $y_i$.


\begin{figure*}[t]
    \centering
    \includegraphics[width=\textwidth]{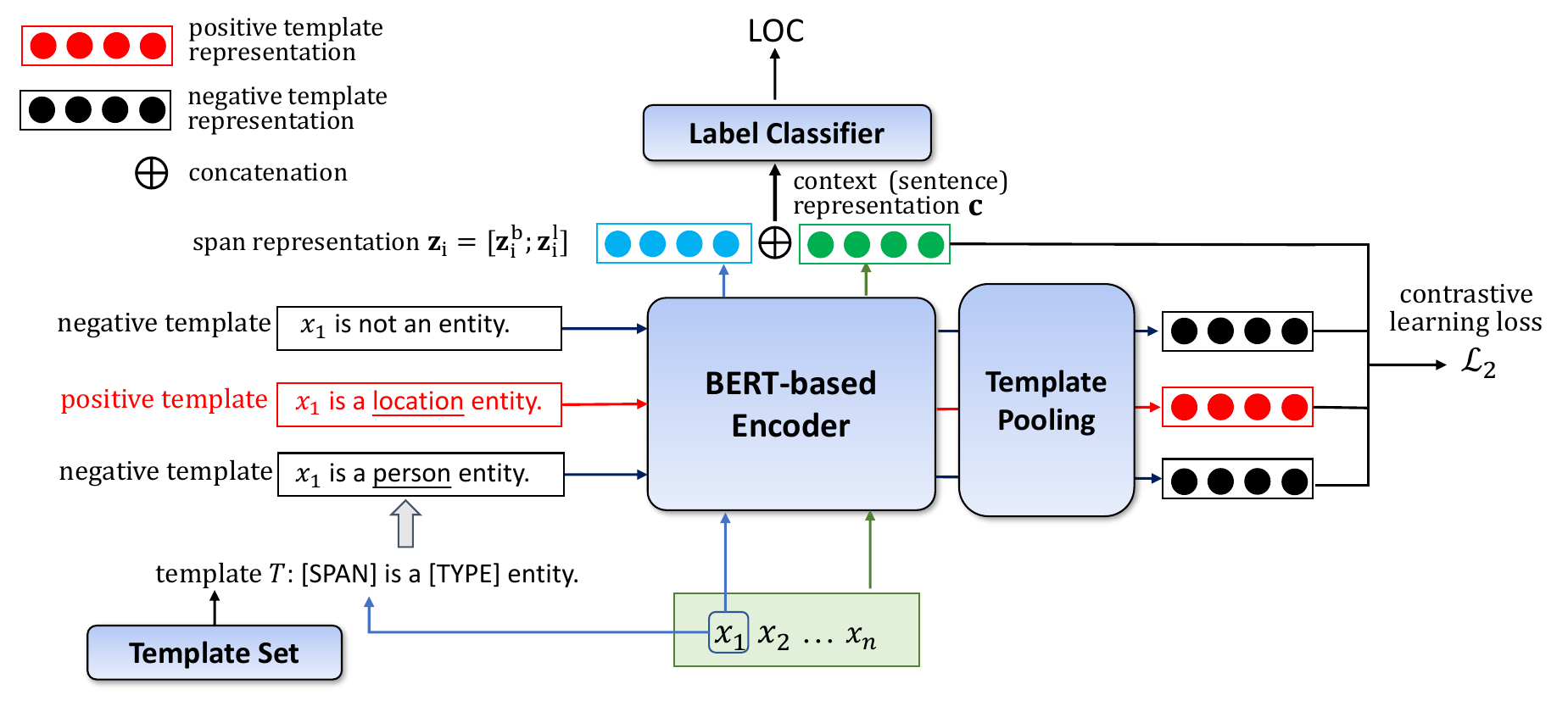}
    \caption{The overall architecture of our proposed S+NER (better viewed in color), which has two major parts: the encoding layer and the label classifier. In the encoding layer fed with the input sentence $X$, the target span's representation $\mathbf{z}_i$ is obtained. In addition, the sentence representation of $X$, denoted as $\mathbf{c}$, is originally generated by the BERT-based encoder, and then refined through the contrastive learning with the positive and negative template representations. Then, $\mathbf{z}_i$ is concatenated with $\mathbf{c}$ and then fed into the classifier to predict the span's label.
    }
    \label{fig:model}
\end{figure*}

\subsection{S+NER}\label{sec:S+NER}
Besides the span's token-level features, we further incorporate the sentence-level information of the span, i.e., the sentence representation, into our framework to achieve NER. Specifically, the sentence representation is further refined by our proposed contrastive learning strategy. In this subsection, we introduce the procedure of generating and refining the sentence representation in our S+NER, of which the overall architecture is shown in Figure \ref{fig:model}.



\subsubsection{Generating Original Context Representation}

As we emphasized in Section \ref{sec:intro}, the contextual information of the mention (span) is significant for a NER model to recognize the entity. An intuitive way to incorporate contextual information is to directly leverage the sentence's representation into which the semantic and syntactic information of the sentence are encoded. Given the power of pre-trained language models (PLMs) such as BERT \cite{devlin2018bert} on understanding sentences, we employ the BERT-based encoder to encode the input sentence into an embedding at first.


Formally, given an input sentence sequence $X = \{x_1, x_2, \cdots, x_n\}$, its original sentence representation is obtained by
\begin{equation}
    \mathbf{c} = \text{SentenceEmbedding}(X) = f(X)\in\mathbb{R}^{d},
\end{equation}
where $\text{SentenceEmbedding}(\cdot)$ and $f(\cdot)$ represents the sentence encoder's operations. In this paper, we use the average pooling of all token representations in a sentence to obtain $\mathbf{c}$.

Then, $\mathbf{c}$ can be used as an auxiliary feature to help the model recognize the target span $s_i$ more accurately. Thus $s_i$'s representation $\mathbf{z}_i$ is expanded with three parts: \textbf{boundary embedding} $\mathbf{z}_i^b$, \textbf{span length embedding} $\mathbf{z}_i^l$ and \textbf{sentence representation} $\mathbf{c}$, that is
\begin{equation}
    \mathbf{z}_i = [\mathbf{z}_i^b; \mathbf{z}_i^l; \mathbf{c}]\in\mathbb{R}^{3d+d'}.
\end{equation}

At last, we feed expanded $\mathbf{z}_i$ into the label classifier which is the same as the span-based NER model's classification layer, to predict the label of $s_i$.

\begin{table*}
\centering
\resizebox{0.62\textwidth}{!}{
\begin{tabular}{c|ccccc}
\toprule\hline
Dataset & Training & Training & Test & Test & OOE \\
  & sen. \# & ent. \# & sen. \# & ent. \# & rate \\ \hline
WNUT2017  & 3,394 & 1,592 & 1,286  & 947 & 100\% \\
JNLPBA   & 18,545 & 18,899  & 3,856 & 4,344 & 67\% \\ 
TwitterNER & 4,000 & 4,899 &  3,257 &  4,106  & 79\%  \\
CoNLL2003-Typos & 14,041 & 8,082 & 2,676 & 4,130 & 86\% \\
CoNLL2003-OOE & 14,041 & 8,082 & 3,453 & 2,689 & 96\% \\
\hline
\bottomrule
\end{tabular}}
\caption{The five datasets' statistics of sentence number and unique entity number in the training and test sets, along with the rate of OOE entities in the test sets.}
\label{table:datasets}
\end{table*}

\subsubsection{Refining Context Representation with Templates}

Specifically, we formally define the \emph{positive/negative template} as follows, which is used in our contrastive learning as the positive/negative golden standard of the context representation. Take an empty template $T = \text{``[SPAN] is a [TYPE] entity.''}$ as an example, where the position of [SPAN] will be filled with the span $s_i$, and the position of [TYPE] will be filled with $s_i$'s correct or incorrect type (label). Thus, we get the $s_i$'s positive template and negative template by filling $T$ respectively with its positive and negative labels. To make the encoder understand the template more effectively, we further translate the label into a term of natural language before filling $T$, e.g., `LOC' is translated into `location'. 

Then, the filled template is also fed into the BERT-based encoder as shown in Figure \ref{fig:model}, to obtain the positive/negative template representation. As we mentioned in Section \ref{sec:intro}, the positive template explicitly indicates the correct type of the entity corresponding to $s_i$, e.g., ``Milan is a location entity.''. Thus, $s_i$'s context representation should be aligned with the representation of this positive golden standard, so as to help the model recognize the entity more accurately. In addition, the negative template also explicitly indicates the misunderstanding of the entity's type, e.g., ``Milan is not an entity.'' or ``Milan is a person entity.''. Thus, $s_i$'s context representation should be simultaneously pushed away from the negative template representation as far as possible.

To leverage the rich semantic information from multiple templates, we employ the state-of-the-art LLM to generate these templates. Specifically, we utilized GPT-4 \cite{gpt-4} to produce 100 templates, which were then filled in as previously described. From these, we manually selected 10 representative templates. Formally, denote $\mathcal{T} = \{T_1, T_2, \cdots, T_k\}$ is the template set. Let $\mathcal{T}_i^+$ and $\mathcal{T}_i^-$ be the positive and negative template sets corresponding to template $T_i$ respectively. After pooling each entity type set along the template dimension, we obtain the template representation for the entire template set. We define the loss function of contrastive learning, inspired by InfoNCE loss \cite{infonce}, as follows:

\begin{align}
    \mathcal{L}_2 = \text{InfoNCE}(\mathbf{c}, \text{Pooling}(\mathcal{T}_i^+), \text{Pooling}(\mathcal{T}_i^-)).
\end{align}

Then, the overall loss of training S+NER is
\begin{equation}
    \mathcal{L} = \mathcal{L}_1 + \lambda \mathcal{L}_2,
\end{equation}
where $\lambda$ is the weight for contrastive learning loss. A larger $\lambda$ value will make the training proportion of the comparative learning part larger, and a smaller $\lambda$ value will make the training proportion of the comparative learning part smaller. A suitable $\lambda$ value will make a good tradeoff between $\mathcal{L}_1$ and $\mathcal{L}_2$, which allows the trained model to make full use of the extracted contextual information to assist in the OOE-NER task.

It is important to highlight that the contrastive learning module is not incorporated during the inference phase of S+NER. This is primarily due to the fact that the true labels of the spans in the test set are not available. Despite the lack of refinement in the context representation of a test sample during model inference, the fine-tuned parameters in S+NER ensure the context representation is adequately optimized. This optimization is crucial as it facilitates accurate prediction, thereby enhancing the overall performance and reliability of the model.
\section{Experiment}
In this section, we display and analyze the results of evaluating our S+NER's OOE-NER performance upon five benchmark OOE datasets, compared with the S+NER NER models. At the same time, we have also conducted ablation experiments on S+NER to verify that each component in the S+NER is reasonable and effective.

\begin{table*}
\centering
\resizebox{1.0\textwidth}{!}{\begin{tabular}{c|ccccc|c}
\toprule\hline
\textbf{Model} & \textbf{WNUT2017}  & \textbf{JNLPBA} & \textbf{TwitterNER} & \textbf{CoNLL2003-Typos} & \textbf{CoNLL2003-OOE} & \textbf{Avg.} \\ \hline
\textbf{BERT-Tagger} & 44.69  & 71.69 & 72.18 & 81.17 & 64.53 & 66.85 \\
\textbf{BERT-CRF} & 43.97 & 72.62 & 73.22 & 81.60  & 65.65 & 67.41 \\
\textbf{DataAug} & 52.29 & 75.85 & 73.69 & 81.73 & 69.60 & 70.63 \\
\textbf{InferNER} & 50.52 & 72.33 & 74.17 & 81.11 & 67.78 & 69.18 \\
\textbf{CoFEE} & 39.10 & 72.55 & 69.50 & 83.13 & 65.44 & 65.94 \\
\textbf{SpanNER} & 51.83 & 73.78 & 71.57& 81.83 & 64.43 & 68.69 \\
\textbf{MINER} & 54.52 & 77.03 & \underline{75.26} & \underline{87.09} & \underline{78.03} & 74.39 \\ 
\textbf{DSpERT} & \underline{55.32} & \underline{81.46} & 74.11 & 85.44 & 77.56 & \underline{74.78} \\
\hline
\textbf{S+NER} & \textbf{58.27}& \textbf{82.70} & \textbf{78.01} & \textbf{90.96} & \textbf{83.52} & \textbf{78.69} \\
\hline
\bottomrule
\end{tabular}}
\caption{Overall performance comparisons between our S+NER and the baselines on the five datasets.}
\label{table:main_results}
\end{table*}

\begin{table}[t]
\centering
\resizebox{0.49\textwidth}{!}{\begin{tabular}{l|l|l}
\toprule\hline
Para. & Value & Comment \\\hline
$d$ & 1,024 & token and sentence representation size\\
$d'$ & 50 & span length embedding size\\
$\tau$ & 1 & the temperature hyperparameter in $\mathcal{L}_2$\\
$\lambda$ & 0.1 & $\mathcal{L}_2$'s weight in the overall loss \\
\hline
\bottomrule
\end{tabular}}
\caption{Some important hyperparameter settings for S+NER implementation.}
\label{tb:set}
\end{table}

\subsection{Datasets and Evaluation Metric}
We conducted our experiments on five datasets including \textbf{WNUT2017} \cite{wnut2017}, \textbf{JNLPBA} \cite{collier2004introduction}, \textbf{TwitterNER} \cite{Zhang_Fu_Liu_Huang_2018}, \textbf{CoNLL2003-Typos} \cite{textflint} and \textbf{CoNLL2003-OOE} \cite{textflint}. 

    
    
    
    

As evaluating previous NER models \cite{fu-etal-2021-spanner,wang2022miner}, we report the entity-level micro F1 scores of all compared models in our experiments. Table \ref{table:datasets} gives some basic information on the training set and the test set of these datasets, including the number of sentences, the number of entities, and the proportion of OOE entities that do not appear in the training set but appear in the test set.

\subsection{Baselines and Implementation Details}
We compared our framework with the baselines including BERT-Tagger \cite{devlin2018bert}, BERT-CRF, SpanNER \cite{fu-etal-2021-spanner}, DataAug \cite{dai2020analysis}, InferNER \cite{shahzad2021inferner}, CoFEE \cite{fukuda-etal-2020-robust}, MINER \cite{wang2022miner} in the experiments. 

The implementation details of our experiments are as follows.
We choose the BERT-large \cite{devlin2018bert} as S+NER's span and sentence encoder in the following comparison experiments. The learning rate for the span classification layer is set to 5e-5, the learning rate for BERT is set to 1e-5, and the dropout rate for the span classification layer is set to 0.2. In order to make a trade-off between the effectiveness and performance of the model, for each input sentence exceeding 128 tokens, we only reserved its first 128 tokens. To limit the number of all extracted spans on the affordable level, we set the maximum length of a span as 4. We conducted our experiments on 1 NVIDIA GeForce RTX 4090 GPU. The checkpoint with the best performance on the validation set will be evaluated on the test set to report the final result. The F1 scores of all experimental results are obtained by averaging the results of five random experiments. Some important hyperparameter settings are listed in Table \ref{tb:set}, most of which were decided based on our tuning studies.

\subsection{Overall Performance Comparisons}
Table \ref{table:main_results} shows the NER performance of our S+NER and all baselines on the five datasets, where the best and second best score in each dataset are bolded and underlined, respectively. In addition, we directly report the scores of InferNER, CoFEE, and MAML provided in the paper of MINER since their source codes have not been published. Except for the JNLPBA dataset, our proposed S+NER outperforms all other baselines on the WNUT2017, TwitterNER, CoNLL2003-Typos, and CoNLL2003-OOE datasets. Compared to the sub-optimal model, S+NER shows a noticeable performance improvement, especially on the WNUT2017 dataset with a 100\% OOE ratio. This demonstrates that S+NER has better robustness in more severe OOE scenarios. Based on the average performance scores across the five datasets, our proposed S+NER achieves the best overall results in OOE-NER. SpanNER serves as the foundation of the span-based NER model, outperforming the two basic sequence tagging models, BERT-Tagger and BERT-CRF. DataAug enhances SpanNER with data augmentation, further improving its OOE-NER capabilities. MINER incorporates the information bottleneck theory into SpanNER to better handle OOE-NER, achieving overall sub-optimal results. By fully leveraging sentence-level information, S+NER surpasses all baselines in OOE-NER, highlighting the crucial role of context information in OOE-NER.

\subsection{Ablation Study}
\begin{table*}[t]
\centering
\resizebox{0.7\textwidth}{!}{
\begin{tabular}{c|lc}
\toprule\hline
\textbf{Datasets}                    & \multicolumn{1}{c}{\textbf{Models}}        & F1 \\ 
\Xhline{1pt}
\multirow{3}{*}{WNUT2017}   & SpanNER                           & 51.83 \\
                            & SpanNER + context representation         & 53.21 \\
                            & SpanNER + template pooling \& contrastive learning & \textbf{58.27} \\ \hline
\multirow{3}{*}{JNLPBA}     & SpanNER                           & 73.78 \\
                            & SpanNER + context representation         & 77.51 \\
                            & SpanNER + template pooling \& contrastive learning & \textbf{82.70} \\ \hline
\multirow{3}{*}{TwitterNER} & SpanNER                           & 71.57 \\
                            & SpanNER + context representation         & 74.48 \\
                            & SpanNER + template pooling \& contrastive learning & \textbf{78.01} \\ 
\hline
\bottomrule
\end{tabular}
}
\caption{Ablation study results on the three datasets.}
\label{table:ablation}
\end{table*}

\begin{table*}[t]
\centering
\resizebox{0.8\textwidth}{!}{
\begin{tabular}{c|c|c}
\toprule\hline
Entity Template   & None-Entity Template & F1 \\ \hline
{[}SPAN{]} is a {[}TYPE{]} entity. &  {[}SPAN{]} is not an entity. & 56.27 \\
{[}SPAN{]} belongs to {[}TYPE{]} category. & {[}SPAN{]} belongs to none category. & 56.92 \\
{[}SPAN{]} should be tagged as {[}TYPE{]}. & {[}SPAN{]} should be tagged as none entity. & 55.75 \\
{[}SPAN{]} can be viewed as {[}TYPE{]} entity. & {[}SPAN{]} can be viewed as none entity. & 54.29 \\ \hline
Template Pooling & Template Pooling & 58.27 \\ 
\hline \bottomrule
\end{tabular}
}\caption{Micro F1 scores of adopting different templates in WNUT2017.}
\label{table:template}
\end{table*}

\begin{table*}[t]
\centering
\resizebox{0.66\textwidth}{!}{
\begin{tabular}{c|c|c|c}
\toprule\hline
\textbf{Model} & \textbf{WNUT2017} & \textbf{JNLPBA} & \textbf{TwitterNER} \\ 
\hline
\textbf{SpanNER(BERT-large)} & 51.83 & 73.78 & 71.57 \\
\textbf{S+NER(BERT-large)} & 58.27 & \textbf{82.70} & 78.01 \\
\hline
\textbf{SpanNER(RoBERTa-large)} & 52.47 & 74.67 & 73.50 \\
\textbf{S+NER(RoBERTa-large)} & \textbf{59.03} & 79.13 & 77.60 \\
\hline
\textbf{SpanNER(DeBERTa-large)} & 52.75 & 74.37 & 73.87 \\
\textbf{S+NER(DeBERTa-large)} & 58.97 & 78.59 & \textbf{79.37} \\ 
\hline\bottomrule
\end{tabular}
}\caption{Micro F1 scores of adopting different encoders in SpanNER and S+NER.}
\label{table:PLM_results}
\end{table*}

To validate the efficacy of S+NER, we conducted an ablation study, the results of which are presented in Table \ref{table:ablation}. In this study, we compared the performance of three methods on the WNUT2017, JNLPBA, and TwitterNER datasets. The first method is the original SpanNER, the second is SpanNER enhanced with context representation, and the third is an extension of the second method, incorporating template pooling and contrastive learning to optimize context representation, which we refer to as S+NER. The ablation study conducted on these three datasets consistently demonstrates that both context representation and contrastive learning, with the templates for context representation, positively impact the performance of the model.

\subsection{Effects of Different Templates}

As previously discussed, S+NER utilizes a template-based contrastive learning and template pooling method. To evaluate the performance of single template on S+NER for OOE-NER, we designed different templates inspired by \citet{template_based} and obtained the results in WNUT2017 shown in Table \ref{table:template}. From the experimental results, it is evident that S+NER is sensitive to template selection, with different templates directly affecting the final OOE-NER performance. The template pooling method, based on multiple templates, concurrently achieved optimal results, which underscores the efficacy of template pooling.

\subsection{Effects of Different Encoders}
We have argued that the special designs in S+NER, i.e., incorporating the context representation and refining it by contrastive learning, are model-agnostic. They can be plugged into the models with a sentence encoder. To verify this characteristic, we compared the variants of SpanNER and S+NER with different pre-trained language models as the encoder, including BERT-large \cite{devlin2018bert}, RoBERTa-large \cite{liu2019roberta} and DeBERTa-large \cite{he2020deberta}. The comparison results shown in Table \ref{table:PLM_results} imply that different pre-trained language models have different capabilities of natural language understanding, resulting in different NER performances. In addition, no matter which pre-trained language model is adopted as the encoder, S+NER consistently outperforms SpanNER, justifying the effectiveness of our special designs.

\subsection{Effects on Different OOE Rates}
We focus on the OOE-NER task in this paper, so we further investigated the compared models' performance in the scenarios with different OOE rates. To this end, we re-partitioned the training and test set to achieve different OOE rates in the dataset, meanwhile keeping the sentence and entity number as close as possible to the original dataset.

\begin{figure}
    \centering
    \includegraphics[width=0.48\textwidth]{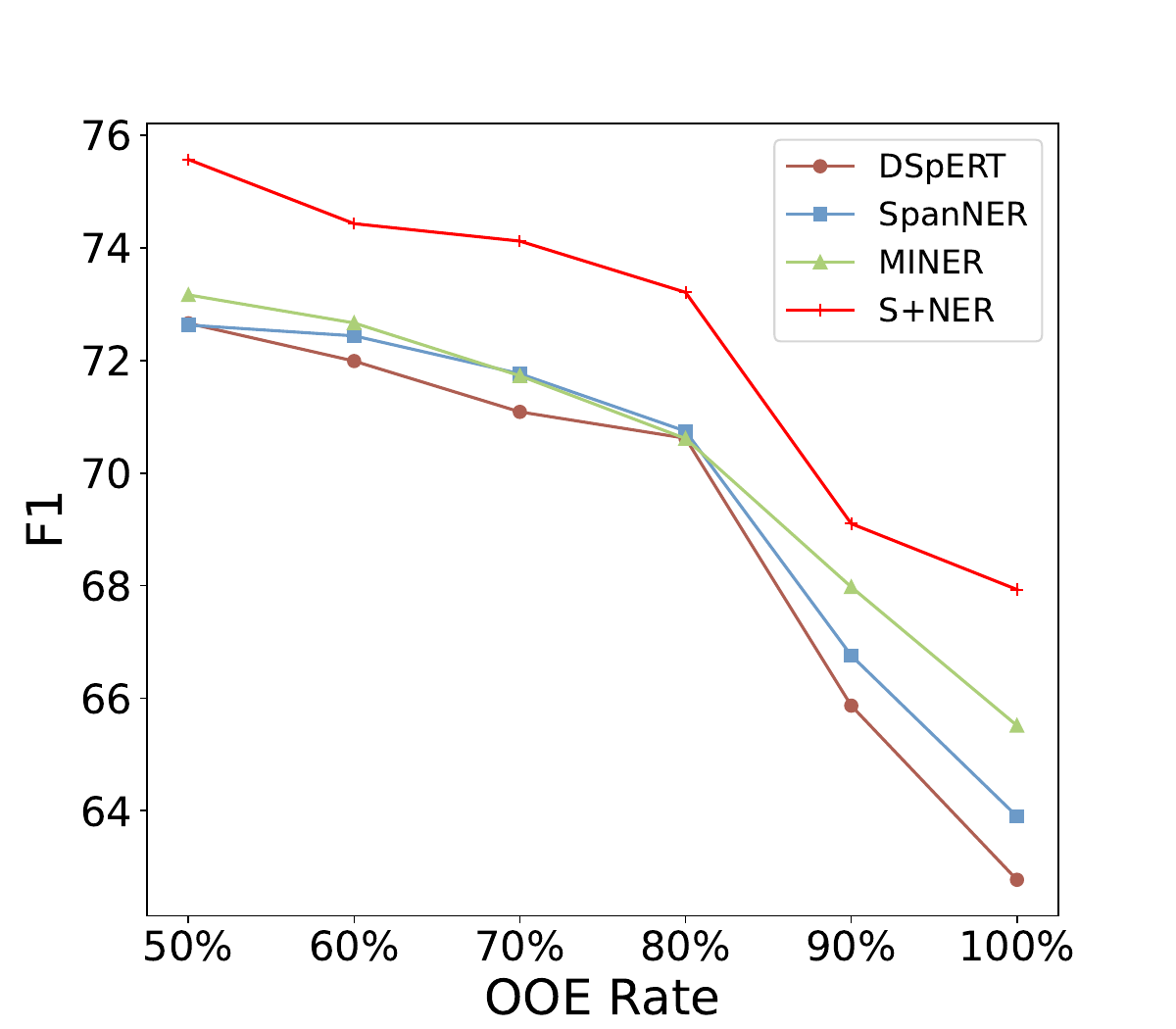}
    \caption{The performance of SpanNER, DSpERT, MINER and S+NER on different OOE rates of TwitterNER. It is worth noting that the TwitterNER dataset here is re-partitioned the training and test set to achieve different OOE rates.}
    \label{fig:OOE_rate_exp}
\end{figure}

We display the results of our S+NER on TwitterNER compared with SpanNER, DSpERT and the previous competitive model MINER in Figure \ref{fig:OOE_rate_exp}, where their performance on six different OOE rates are displayed. It shows that, although all models' performance degrades as the OOE rate increases, S+NER outperforms the three baselines on all OOE rates consistently. This result highlights the benefits of our proposed S+NER in OOE-NER compared to previous models. As the OOE rate increases (>80\%), all models exhibit a significant drop in performance, underscoring the severity of the OOE issue.

\section{Conclusion}
In this paper, we propose a novel framework S+NER to handle Out-of-Entity Named Entity Recognition, in which we have two major designs: incorporating the sentence representation of the input sequence and then refining it with a contrastive learning and template pooling strategy. These two designs can help our framework better understand the contextual information of the target entity, and thus alleviate the OOE problem for achieving better OOE-NER. Our extensive experiments demonstrate our S+NER's advantage over the SOTA OOE-NER models. In addition, we also conduct experiments to examine the impact of various templates, pre-trained language model encoders, and OOE rates on the performance of S+NER. Through case studies, we intuitively demonstrate that S+NER outperforms other models.

\section*{Limitations}
As we analyzed in the experiment section, our proposed framework relies on the PLM's capability of understanding the span's context. For the data from a professional field or a special domain, a PLM may not well understand the contexts since it is generally pre-trained with the corpus of open domains. In such scenario, re-pre-training the PLM or fine-tuning with the data from the special domain is expected for achieving satisfactory NER performance.

\section*{Acknowledgements}
This work was supported by the Chinese NSF Major Research Plan (No.92270121).

\bibliography{custom}


\end{document}